\documentclass{article}
\usepackage{ijcai21}
\usepackage{times}

\usepackage{soul}
\usepackage{url}
\usepackage[hidelinks]{hyperref}
\usepackage[utf8]{inputenc}
\usepackage[small]{caption}
\usepackage{graphicx}
\usepackage{amsmath}
\usepackage{booktabs}
\usepackage{caption}
\usepackage{subcaption}
\usepackage{helvet} 
\usepackage{amssymb} 
\usepackage{comment} 
\usepackage{booktabs, multirow} 
\usepackage{soul}
\usepackage[table]{xcolor} 
\usepackage{changepage,threeparttable} 

\usepackage{subcaption}
\usepackage{xcolor}
\usepackage{amsfonts,amssymb}
\usepackage[ruled,vlined,linesnumbered]{algorithm2e} 
\usepackage{enumitem}

\SetKwInput{KwInput}{Input}                
\SetKwInput{KwOutput}{Output}              
\SetKwInput{KwRequire}{Require}              
\newenvironment{packed_item}{
	\begin{itemize}[leftmargin=*,topsep=0pt,partopsep=0pt]
		\setlength{\itemsep}{0pt}
		\setlength{\parskip}{0pt}
		\setlength{\parsep}{0pt}
	}
	{\end{itemize}}

\newenvironment{packed_enum}{
	\begin{enumerate}[leftmargin=*,topsep=0pt,partopsep=0pt]
		\setlength{\itemsep}{0pt}
		\setlength{\parskip}{0pt}
		\setlength{\parsep}{0pt}
		\setlength{\topsep}{0pt}
	}
	{\end{enumerate}}

\newcommand{\kmethod}{AQ-BERT }

\title{Automatic Mixed-Precision Quantization Search of BERT}
\author{
Changsheng Zhao\and
Ting Hua\and
Yilin Shen\and
Qian Lou \and
Hongxia Jin
\affiliations
Samsung Research America
\emails
\{changsheng.z, ting.hua, yilin.shen, qian.lou, hongxia.jin\}@samsung.com
}

\begin{document}

\maketitle
\begin{abstract}
Pre-trained language models such as BERT have shown remarkable effectiveness in various natural language processing tasks. 
However, these models usually contain millions of parameters, which prevents them from  practical deployment on resource-constrained devices.  
 \textit{Knowledge distillation},  \textit{Weight pruning}, and   \textit{Quantization} are known to be the main directions in  model compression. 
However, compact models obtained through {knowledge distillation}  may suffer from significant accuracy drop even for a relatively small compression ratio. 
On the other hand, there are only a few {quantization} attempts that are specifically designed for natural language processing tasks. They suffer from a small compression ratio or a large error rate since manual setting on hyper-parameters is required and fine-grained subgroup-wise {quantization} is not supported.  
In this paper, 
we proposed an automatic mixed-precision quantization framework designed for BERT that can simultaneously conduct quantization and pruning in a subgroup-wise level.
Specifically, our proposed method leverages \textit{Differentiable Neural Architecture Search} to assign scale and precision for parameters in each sub-group automatically, and at the same time pruning out redundant groups of parameters. 
Extensive evaluations on BERT downstream tasks reveal that our proposed method outperforms baselines by providing the same performance with much smaller model size.
We also show the feasibility of obtaining the extremely light-weight model by combining our solution with orthogonal methods such as DistilBERT.
\end{abstract}

\section{Introduction}
Transformer based architectures such as  BERT ~\cite{devlin2019bert}, have achieved significant performance improvements over traditional models in a variety of Natural Language Processing tasks.
These models usually require long inference time and huge model size with million parameters despite their success.
For example, an inference of BERT$_{Base}$ model involves $110$ million parameters and $29$G floating-point operations.
Due to these limitations, it is impractical to deploy such huge models on resource-constrained devices with tight power budgets.

\textit{Knowledge distillation}, \textit{Pruning}, and  \textit{Quantization} are known to be three promising directions to achieve model compression. 
Although compression technologies have been applied to a wide range of computer vision tasks,
they haven't been fully studied in natural language processing tasks.
Due to the high computation complexity of pre-trained language models, it is nontrivial to explore the compression of  transformer-based architectures.

\textit{Knowledge distillation} is the most popular approach in the field of pre-trained language model compression.
Most current work in this direction usually reproduces the behavior of a larger teacher model into a smaller lightweight student model \cite{sun2019patient,sanh2019distilbert,jiao2019tinybert}.
However, the compression efficiency of these models  is still low and
significant performance degradation is observed even at a relatively small compression ratio.
\textit{Pruning} a neural network means removing some neurons within a group.
After the pruning process, these neurons' weights are all zeros, which decreases the memory consumption of the model.
Several approaches have explored weight pruning for Transformers \cite{gordon2020compressing,kovaleva2019revealing,michel2019sixteen}, which mainly focus on identifying the pruning sensitivity of different parts.
\textit{Quantization} is a model-agnostic approach that is able to reduce memory usage and improve inference speed at the same time.
Compared to {Knowledge distillation} and {Pruning}, there are much fewer attempts based on quantization for Transformer compression.
The state-of-the-art quantization method Q-BERT adapts \textit{mixed-precision} quantization for different layers \cite{shen2019q}.
However, Q-BERT determines the quantization bit for each layer by hand-crafted heuristics.
As BERT-based models become deeper and more complex, the design space for \textit{mixed-precision} quantization increases exponentially, which is challenging to be solved by heuristic, hand-crafted methods.

In this paper, we proposed an automatic mixed-precision quantization approach for BERT compression (AQ-BERT).
Beyond layer-level quantization, our solution is a group-wise quantization scheme.
Within each layer, our method can automatically set different scales and precision for each neuron sub-groups.  
Unlike Q-BERT that requires a manual setting, we utilize differentiable network architecture searches to make the precision assignments without additional human effort. 
Our contributions can be summarized as follows.



    \paragraph{Proposal of an united framework to achieve automatic parameter search.}
    Unlike existing approaches, \kmethod does not need hand-craft adjustments for different model size requirements.
    This is achieved by designing a two-level network to relax the precision assignment to be continuous, which can therefore be optimized by gradient descent.
    \paragraph{Proposal of a novel objective function that can compress model to the desirable size, and enables pruning and quantization simultaneously.}
    Given a targeted model size, our \kmethod aims to search for the optimal parameters, regularized by minimizing the cost computation.
    As the the cost is a group Lasso regularizer, this design also enables a joint pruning and quantization.
    
    \paragraph{Provide efficient solutions to optimize the parameters for the proposed framework.}
    The optimization towards the objective of the proposed framework is non-trivial, as the operations for quantization are non-differentiable.
    
    \paragraph{Extensive experimental validation on various NLP tasks.}
    We evaluate the proposed \kmethod on four NLP tasks, including Sentiment Classification, Question answer, Natural Language Inference, and Named Entity Recognition.
    The results demonstrate that our \kmethod achieves superior performance than the state-of-the-art method Q-BERT.
    And we also show that the orthogonal methods based on knowledge distillation can further reduce the model size. 

\section{Related Work}
Our focus is model compression for Transformer encoders such as BERT.
In this section, we first discussed the main  branches of  compression  technologies for general purpose, then we reviewed existing work specifically designed for compressing Transformers.
Besides compression technologies, our proposed method is also related to the filed of 
Network architecture search.

\subsection{General Model Compression}
\label{sec:related_general}
A traditional understanding is that a large number of parameters is necessary for training good deep networks \cite{zhai2016doubly}. However, it has been shown that many of the parameters in a trained network are redundant \cite{han2015deep}. 
Many efforts have been made to model compression, in order to deploy efficient models on the resource-constrained hardware device.

Automatic network pruning is one promising direction for model compression, which removes unnecessary network connections to decrease the complexity of the network. 
This direction can be further divided into pruning through regularization and network architecture search.
Pruning based on regularization usually adds a heuristic regularizer as the penalty on the loss function \cite{molchanov2017variational,louizos2018learning}.
While pruning through network architecture search aims to discover the important topology structure of a given network \cite{dong2019network,frankle2018lottery}.

Another common strategy is weight quantization, which constrains weight to a set of discrete values.
By representing the weights with fewer bits, quantization approaches can reduce the storage space and speed up the inference. 
Most of the quantization work assign the same precision for all layers of a network \cite{rastegari2016xnor,choi2018pact}.
And the few attempts on mixed-precision quantization are usually on layer-level \cite{zhou2018adaptive,wu2018mixed}, without support for the assignments on the finer level such as sub-groups.


Besides, knowledge distillation is also a popular direction for model compression \cite{hinton2015distilling}, which learns a compact model (the student) by  imitating the behavior of a larger model (the teacher).

\subsection{Transformers Compression}

Existing attempts on compressing Transformers are mainly based on knowledge distillation \cite{sanh2019distilbert,liu2019multi,jiao2019tinybert}, which is the orthogonal direction to our solution. 

Most technologies based on pruning and quantization mentioned above, are deployed to convolutional networks, while only a few works are designed for deep language models such as Transformers. 
And the majority of these work focus on heuristic pruning \cite{kovaleva2019revealing,michel2019sixteen}  or 
study effects of pruning at different levels \cite{gordon2020compressing}. 
Q-BERT  is most related to our work that is also a mixed-precision quantization approach designed for BERT \cite{shen2019q}.
However, they require extensive efforts to manually set the hyper-parameters, which is infeasible for practical usage.

\subsection {Network Architecture Search}
The problem of network compression can also be viewed as a type of sparse architecture search.
Although most previous research on network architecture search \cite{zoph2016neural,hu2020dsnas} can automatically discover the topology structure of deep neural networks,
they usually require huge computational resources.
To reduce the computational cost, ENAS \cite{pham2018efficient} shares the weights of a super network to its child network, and utilizes reinforcement learning to train a controller to sample better child networks. 
DARTS \cite{liu2018darts}  also is a two-stage NAS that leverages the differentiable loss to update the  gradients.
We inherit the idea of a super network from them to the field of model compression, by simultaneously conducting quantization and pruning during the architecture search process.

\section{Methods}


\label{sec:model}


\begin{figure*}[htbp]
\centering
\includegraphics[width=\textwidth]{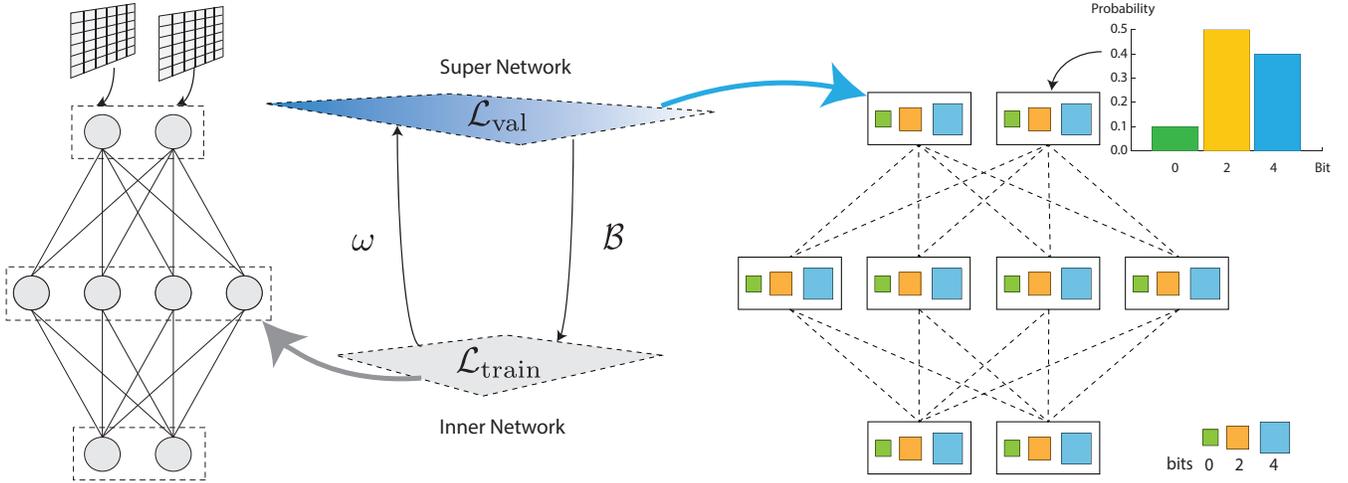}
\caption{Framework. The central part shows the idea of two-stage optimization.
The left part is the illustration of the inner training network, while the right part is an example of the super network that controls bit-assignment.
In the left part, each node represents a matrix (a group of neurons), which we call a ``sub-group'' in our paper. Each sub-group has its own quantization range in the mixed-precision setting. As the example shown in the right part, a sub-group has three choices of bit-assignment: 0-bit, 2-bit, and 4-bit. And each such assignment is associated with a probability of being selected.
}
\label{fig:framework}
\end{figure*}

Given a large model $\mathcal{M}$, our goal is to obtain a compact model $\mathcal{M}'$ with desirable size $\mathcal{V}$, by automatically learning the optimal bit-assignment set $\mathcal{O}^*$ and weight set $\omega^*$. 
To achieve this goal, we have to solve the following challenges:
\begin{packed_enum}
    \item How to find the best bit assignment automatically?
    \item Is it possible to achieve pruning and quantization simultaneously?
\item How to compress the model to a desirable size?
    \item Bit assignments are discrete operations, how to achieve back propagation under this condition?
    \item How to efficiently infer the parameters for  bit-assignment set and weight set together?
\end{packed_enum}
This section discusses the solutions to all these challenges. 
Our proposed framework makes it possible to automatically search for the best assignment.
Through the carefully designed loss function, a large model can be reduced to a compact model with pre-given desirable size via a compressing process that can conduct pruning and quantization at the same time. 
We show how to make the whole process differentiable through the description of quantization and continuous relaxation. 
Finally, we discuss the optimization process and provide an overall algorithm.

\subsection{Framework}

Figure \ref{fig:framework} illustrates our key ideas.
Specifically, our framework includes  two  networks: a weight-training inner network (the left part),
and a bit-assignment super network (the right part).

The weight-training inner network can be viewed as a regular neural network that optimizing weights, except that each node represents a subgroup of neurons (the slices of dense) rather than a single neuron.

As shown in the right part of  Figure \ref{fig:framework},
for a subgroup $j$ in layer $i$, there could be $K$ different choices of precision, and  the $k$-th choice is denoted as $b_k^{i,j}$ (e.g., 2 bit).
For example,  in Figure \ref{fig:framework}, each subgroup has 3 choices of bit-width: 0 bit, 2-bit, and 4-bit.
Correspondingly, the probability of choosing a certain precision is denoted as $p_k^{i,j}$, and the bit assignment is a one-hot  variable $O_k^{i,j}$.
It is obvious that $\sum_k p_k^{i,j}=1$ and only one precision is selected at a time.

Remember that our goal is achieved by jointly learning the bit assignments  $\mathcal{O}$ and the weights $\omega$ within all the mixed operations.
The super network will update the bit assignment set $\mathcal{O}$ by calculating the validation loss function $\mathcal{L}_{val}$.
And inner training network will optimize weights set $\omega$ through loss function $\mathcal{L}_{train}$ based on cross-entropy. 
We will introduce the details of the loss function in the following section. 

As can be seen from the above discussion, the two-stage optimization framework enables the automatic search for the bit-assignment, which is originally manually set up in Q-BERT.



\subsection{Objective Function}





As stated above, we will jointly optimize the bit-assignment set 
$\mathcal{O}$ and weight set $\omega$.
Both validation loss $\mathcal{L}_{val}$ and training loss 
$\mathcal{L}_{train}$ are  determined not only by the bit assignment $O$, but also the weights $\omega$ in the network. 
The goal for bit-assignment search is to find the best $O^*$ that minimizes the validation loss $\mathcal{L}_{val}(\omega^*_\mathcal{O},\mathcal{O})$, where optimal weight set $\mathcal{\omega}^*$ associated with the bit assignments are obtained by minimizing the training loss $\mathcal{L}_{train}(\mathcal{O}^*,\mathcal{\omega})$.
This is a two-level optimization problem that bit-assignment set $O$ is upper-level variable, and weight set $\omega$ is lower-level variable:

\begin{equation}
    \min_\mathcal{O} \quad \mathcal{L}_{\rm{val}} (\omega^*,\mathcal{O}) \\
\label{equ:obj1}
\end{equation}
\begin{equation}
  s.t. \quad
 \omega^* = \arg \min_\omega
  \mathcal{L}_{\rm{train}}(\omega,\mathcal{O}) 
  \label{equ:obj2}
\end{equation}


The training loss $\mathcal{L}_{train}(\mathcal{O}^*,\mathcal{\omega})$ is a regular cross-entropy loss.
The validation loss $\mathcal{L}_{\rm{val}}$ contains both classification loss and the penalty for the model size:

\begin{equation}
    \mathcal{L}_{\rm{val}} =  -\log{\frac{\exp(\psi_y)}{\sum^{|\psi|}_{j=1}\exp(\psi_y)}}  + \lambda \mathcal{L}_{\rm{size}} 
\label{equ:loss_val}
\end{equation}


$\psi_y$ is the output logits of the network, where $y$ is the ground truth class, and $\lambda$ is the weight of penalty.
We can configure the model size through the penalty  $\mathcal{L}_{\rm{size}}$, which encourages the computational cost of the network to converge to a desirable size $\mathcal{V}$. 
Specifically, the computation cost $\mathcal{L}_{\rm{size}}$ is calculated as follows:

\begin{equation}
\mathcal{L}_{\rm{size}}=
\begin{cases}
log \mathbb{E}[C_{\mathcal{O}}]& C_{\mathcal{O}} > (1+\epsilon)\times \mathcal{V}\\
0& C_{\mathcal{O}} \in [(1-\epsilon)\times \mathcal{V}, (1+\epsilon)\times \mathcal{V}]\\
-log \mathbb{E}[C_{\mathcal{O}}]& C_{\mathcal{O}} < (1-\epsilon)\times \mathcal{V}
\end{cases}
\label{equ:loss_size}
\end{equation}

\begin{equation}
C_{\mathcal{O}}=\sum_{i,j}\sum_{k}||b_k^{i,j}\cdot O_k^{i,j}||_2
\label{equ:C_o}
\end{equation}

\begin{equation}
\mathbb{E}[C_{\mathcal{O}}]=\sum_{i,j}\sum_{k} p_k^{i,j} ||b_k^{i,j}\cdot O_k^{i,j}||_2
\end{equation}
$C_{\mathcal{O}}$ is the actual size of the model with bit-assignment $\mathcal{O}$, which is a group Lasso regularizer.
For a sub-group $j$ on layer $i$, there is a possibility that its optimal bit-assignment is zero. 
In this case, the bit-assignment is equal to pruning that removes this sub-group of neurons from the network. 
Toleration rate $\epsilon \in [0, 1]$ restricts the variation of model size is around the desirable size $\mathcal{V}$. 
$\mathbb{E}[C_{\mathcal{O}}]$ is the expectation of the size cost $C_{\mathcal{O}}$, where the weight is the bit-assignment probability.

The statement and analysis above show the carefully designed validation loss  $\mathcal{L}_{\rm{val}}$ provides a solution to solve the first and the second challenge simultaneously. 
Specifically, it can configure the model size according to the user-specified value $\mathcal{V}$ through piece-wise cost computation, and provide a possibility to achieve quantization and pruning together via group Lasso regularizer.

\begin{table*}[!htp]\centering
\begin{tabular}{l|r|ccccccc}\toprule
Model &Size/MB &SST-2(Acc) &MNLI-m (Acc) &MNLI-mm (Acc) & SQuAD(EM) &SQuAD(F1) &CoNLL(F1) \\\toprule
BERT$_{base}$ &324.5 &93.50 &84.00 &84.40 & 81.54 & 88.69 & 95.00\\\midrule
Q-BERT &30 &92.50 &83.50 &83.50 &79.07 &87.49 &94.55 \\
Ours &30 &92.70 &83.50 &83.70 &79.85 &87.00 &94.50 \\\midrule
Q-BERT &25 &92.00 &81.75 &82.20 &79.00 &86.95 &94.37 \\
Ours &25 &92.50 &82.90 &82.90 &79.25 &87.00 &94.40 \\\midrule
Q-BERT &20 &84.60 &76.50 &77.00 &69.68 &79.60 &91.06 \\
Ours &20 &\textbf{91.10} &\textbf{81.80} &\textbf{81.90} &\textbf{75.00} &\textbf{83.50} &\textbf{93.20} \\
\bottomrule
\end{tabular}
\caption{Quantization results of Q-BERT and our method for BERT$_{base}$ on Natural Language Understanding tasks. Results are obtained with 128 groups in each layer.   Both Q-BERT and our method are using 8-bits activation. All model sizes reported here exclude the embedding layer, as we uniformly quantized embedding by 8-bit.}
\label{tab:q_main}
\end{table*}

\subsection{Quantization Process}
Traditionally, all the weights in a neural network are represented by full-precision floating point numbers (32-bit).
Quantization is a process that converts full-precision weights to fixed-point numbers with lower bit-width, such as 2,4,8 bits.
In mixed-precision quantization, different groups of neurons can be represented by different quantization range (number of bits).


If we denote the original floating-point sub-group in the network by matrix $A$, and the number of bits used for quantization by $b$, then we can calculate its own scale factor  $q_A \in \mathbb{R}^+$ as follows:

\begin{equation}
    q_A=\frac{2^b-1}{max(A)-min(A)}.
    \label{equ:scaler}\\
\end{equation}
And a floating-point element $a \in A$ that can therefore be estimated  by the scale factor and its quantizer $Q(a)$ such that $a \approx   Q(a)/q_A$.
Similar to Q-BERT, the uniform quantization function is used to evenly split the range of floating point tensor \cite{hubara2017quantized}:
\begin{equation}
    Q(a)=round(q_A\cdot[a-min(A)]).
    \label{equ:q_a}
\end{equation}

The quantization function is non-differentiable,
therefore the ``Straight-through estimator'' (STE)
 method  is needed here to  back-propogate the gradient \cite{bengio2013estimating}, which can 
be viewed as an operator that has arbitrary forward and backward operations:

\begin{align}
    \rm{Forward}: \hat{\omega}_A=Q(\omega_A)/q_{\omega_A} \label{equ:forward}\\
    \rm{Backward}: \frac{\partial \mathcal{L}_{\rm{train}}}{\partial \hat{\omega}_A}=\frac{\partial \mathcal{L}_{\rm{train}}}{\partial \omega_A}.
    \label{equ:quant}
\end{align}
Specifically, the real-value weights $\omega_A$ are converted into the fake quantized weights $\hat{\omega}$ during forward pass, calculated via Equation \ref{equ:scaler} and \ref{equ:q_a}. 
And in the backward pass, we use the gradient $\hat{\omega}$ to approximate the true gradient of $\omega$ by STE.

\subsection{Continuous Relaxation} 

Another challenge is that mixed-precision assignment operations are discrete variables, which are non-differentiable and therefore unable to be optimized through gradient descent. 
In this paper, we use concrete distribution to relax the discrete assignments by using Gumbel-softmax:
\begin{equation}
\begin{split}
    {O}_k^{i,j} = \frac{\exp ((\log \beta_k^{i,j} + g_k^{i,j})/t)}{\sum_k\exp  ((\log \beta_k^{i,j} + g_k^{i,j})/t)} \\
    s.t. \quad g_k^{i,j}=-log(-log(u)), u \sim U(0,1)
    \end{split}
\label{equ:o_k}
\end{equation}
$t$ is the softmax temperature that controls the samples of Gumbel-softmax. 
As $t \rightarrow \infty$, ${O}_k^{i,j}$ is close to a continuous variable following a uniform distribution, 
while $t \rightarrow 0$, the values of ${O}_k^{i,j}$ tends to be one-shot variable following the categorical distribution. 
In our implementation, an exponential-decaying schedule is used for annealing the temperature:
\begin{equation}
    t = t_0 \cdot \rm{exp}(-\eta \times (epoch-N_0)),
\label{equ:temp}
\end{equation}
where $t_0$ is the initial temperature, $N_0$ is the number of warm up epoch, and the current temperature decays exponentially after each epoch.

The utilization of the Gumbel Softmax trick effectively renders our proposed \kmethod  into a differentiable version.

\subsection{Optimization Process}
The optimizations of the two-level variables are non-trivial due to a large amount of computation. 
One common solution is to optimize them alternately, that  
the algorithm infers one set of parameters while fixing the other set of parameters. 
Previous work usually trains the two levels of variables separately \cite{xie2018snas}, which is computationally expensive.
We adopted a faster inference that can simultaneously learn variables of different level \cite{liu2018darts,luketina2016scalable}.

In this paper, validation loss $\mathcal{L}_{\rm{val}}$ is determined by both the lower-level variable weight $\omega$ and the upper-level variable bit assignments $\mathcal{O}$.

\begin{align}
    &\nabla_\mathcal{O}\mathcal{L}_{\rm{val}} (\omega^*,\mathcal{O})\\
\approx & \nabla_\mathcal{O}\mathcal{L}_{\rm{val}} (\omega - \xi \nabla_\omega \mathcal{L}_{\rm{train}},\mathcal{O}) \label{equ:loss_app_1}
\end{align}
It is generally believed that hyper-parameter set $\mathcal{O}$ should be kept fixed during the training process of inner optimization (Equation \ref{equ:obj2}).
However, this hypothesis is shown to be  unnecessary that it is possible to change hyper-parameter set during the training of inner optimization \cite{luketina2016scalable}.
Specifically, as shown in Equation \ref{equ:loss_app_1}, the approximation $\omega^*$ is achieved by adapting one single training step $\omega - \xi \nabla_\omega \mathcal{L}_{\rm{train}}$. 
If the inner optimization already reaches a local optimum ($\nabla_\omega \mathcal{L}_{\rm{train}} \rightarrow 0$), 
then Equation \ref{equ:loss_app_1} can be further reduced to  $\nabla_\mathcal{O} \mathcal{L}_{\rm{val}}(\omega,\mathcal{O})$.
Although the convergence is not guaranteed in theory \cite{luketina2016scalable}, we observe that the optimization process is able to reach a fixed point in practice.
The details can be found in the supplementary material. 


\subsection{Overall Algorithm}

\begin{algorithm}[t]\small
\DontPrintSemicolon
  \KwInput{training set $\mathbb{D}_{train}$ and validation set $\mathbb{D}_{val}$}
  \For {epoch=0,...,N}
  {
  get current temperature via  Equation \ref{equ:temp}\;
  Calculate $ \mathcal{L}_{\rm{train}}$ on $\mathbb{D}_{train}$ to update weights $\omega$\;
  \If{$epoch > N_1$}
  {
  Calculate $ \mathcal{L}_{\rm{val}}$ on $\mathbb{D}_{val}$   via Equation \ref{equ:loss_app_1} to update bit assignments $\mathcal{O}$\;
  }
  }
 Derive the final weights based on learned optimal bit assignments $\mathcal{O}^*$\;
  \KwOutput{ optimal bit assignments $\mathcal{O}^*$ and weights $\omega*$}
\caption{The Procedure of \kmethod}
\label{alg:pipeline}
\end{algorithm}

Based on the statements above, Algorithm \ref{alg:pipeline} summarizes the overall process of our proposed method.
\begin{packed_enum}
    \item As shown in line 2 Algorithm \ref{alg:pipeline}, in the beginning of each epoch,
the bit-assignment is relaxed to continuous variables via Equation \ref{equ:o_k}, where the temperature is calculated through Equation \ref{equ:temp}.
After this step, both weight and  bit-assignment are differentiable. 
\item Then, $ \mathcal{L}_{\rm{train}}$ is minimized on the training set to optimize the weights (line 3).
\item To ensure the weights are sufficiently trained before the updating of the bit assignments, 
we delay the training of $ \mathcal{L}_{\rm{val}}$ on the validation set for $N_1$ epochs (line 4 and 5).
For each sub-group, the number of bits with maximum probability is chosen as its bit assignment.
\item After sufficient epochs, we are supposed to obtain a set of bit assignment that is close to the optimal. Based on current assignments, we then randomly initialize the weights of the inner network, and train it from scratch (line 6).
\end{packed_enum}

With these steps, we can obtain the outputs of the whole learning procedure, which contains the optimized bit assignments and weight matrices.

\section{Experimental Evaluation}
\label{sec:exp}

In this section, we evaluate the proposed \kmethod from the following aspects:
\begin{packed_enum}
    \item How does \kmethod perform comparing to state-of-the-art BERT compression method based on quantization (e.g., Q-BERT)? 
\item How do the parameter settings affect the performance of \kmethod? 
\item Is it possible to integrate the proposed \kmethod with an orthogonal method based on knowledge distillation? 
\end{packed_enum}

To answer these questions, we first compare our \kmethod with baseline under different constraints of model size, then we study the effect of group numbers towards performance, and finally, we present the results of integrating our method with the knowledge distillation approach.

\subsection{Datasets and Settings}
We evaluate our proposed \kmethod and other baselines (bert-base, Q-BERT, and Distilbert-base) on four NLP tasks: SST-2,  MNLI,  CoNLL-2003, and  SQuAD.
 Our implementation is based on transformers by huggingface\footnote{https://github.com/huggingface/transformers}.
The AdamW optimizer is set with learning rate $2e-5$, and SGD is set with  learning rate 0.1 for architecture optimization.

\subsection{Main Results}
\label{sec:q-bert}
In this section, we report the results of comparing our proposed method with baselines on the development set of the four tasks: SST-2, MNLI, CoNLL-03, and SQuAD.  
\thesubsubsection{Performance on different sizes}

Table \ref{tab:q_main} compares our results with Q-BERT on four NLP tasks: SST-2, MNLI, CoNLL-03, and SQuAD.
Several observations can be made from the table as follows:
\paragraph{Overall comparison.}
    As can be seen from the table, in all four tasks,  \kmethod generally performs better than Q-BERT, regardless of the compressed model size. 
    And the performance gap between our method and Q-BERT becomes more obvious as the model size decreases. 
    These observations indicate that, compared to Q-BERT, our proposed \kmethod can learn parameters correctly and perform stably on various tasks. 
\paragraph{Obvious advantage in ultra-low bit setting.}
    Our advantage is more obvious for ultra-low bit setting.
Specifically, when the model size is as small as 20M, \kmethod achieves significant improvements over Q-BERT, as measured by the difference on development set scores for four representative NLP tasks: SST-2 (+6.5\%),  MNLI (+5.3\%), SQuAD (+5.3\%), CoNLL (+2.1\%).
    This phenomenon further confirms the effectiveness of our automatic parameter search.
    As the model size decreases, the optimal set of parameters becomes ``tighter''.
    In this situation, it is less likely to find good settings for parameters through manual assignment adopted by baseline Q-BERT.

\subsubsection{Effects of Group-wise Quantization}

\begin{table}[!tbp]\centering

\small
\begin{tabular}{c|c|p{1cm}p{0.9cm}p{0.9cm}p{0.9cm}}\toprule
Model &Group &SST-2 &MNLI-m &MNLI-mm &CONLL \\\toprule
BERT$_{base}$& N/A & 93.00 & 84.00&84.40 & 95.00\\
\midrule
Q-BERT& 1 & 85.67 & 76.69&77.00&89.86 \\
Ours& 1 &89.60 &77.70&78.20 &91.90 \\\midrule
Q-BERT& 12&92.31&82.37&82.95&94.42\\
Ours& 12 &92.70 &83.50&83.70 &94.80 \\\midrule
Q-BERT& 128&92.66&83.89&84.17&94.90\\
Ours& 128 &92.90 &83.40&83.90 &95.00 \\\midrule
Q-BERT& 768&92.78&84.00&84.20&94.99\\
Ours& 768 &92.90 &83.70&84.10 &95.10 \\
\bottomrule
\end{tabular}
\caption{Effects of group-wise quantization for AQ-BERT. The quantization bits were set to be 8 for embeddings and activations on all the tasks. From top to down, we increase the number of groups. 
Notice that Q-BERT reports the group-wise quantization performance with a model of 40M. To make a fair comparison, we add the choice of 8-bit in this experiment to produce a model with a comparable size. 
}
\label{tab:group}
\end{table}

\begin{table}[!tbp]\centering

\small

\begin{tabular}{p{1.4cm}|c|p{0.6cm}p{1cm}p{1cm}p{1cm}p{1cm}}\toprule
Model & Size/MB & SST-2(Acc) &MNLI-m &SQuAD (EM) &SQuAD (F1) \\\midrule
BERT &324.5 &93.50 &84.00 &81.54 &88.69 \\\midrule
DistilBERT &162 &91.30 &82.20 &77.70 &85.80 \\\midrule
 {} &15 &91.20 &81.50 &72.80 &82.10 \\
DistilBERT &12.5 &90.70 &80.00 &72.70 &82.10 \\
{+Ours} &10 &89.70 &78.00 &68.30 &78.30 \\
\bottomrule
\end{tabular}
\caption{Results of combining our proposed \kmethod with knowledge distillation model DistilBERT.
All the model sizes reported in this table exclude the embedding layer, as we uniformly quantized embedding to 8-bit.}
\label{tab:distil}
\end{table}

 Table \ref{tab:group} shows the performance gains with different group numbers. 
 A larger group number means each sub-group in the network contains fewer neurons.
 For example, in the setting of the 12-group, each group contains $768/12=64$ neurons, while in the setting of the 128-group, each group only contains six neurons.
 Several observations can be made from Table \ref{tab:group} as follows:
 \paragraph{Overall comparison.}
 As can be seen from the table, in terms of accuracy, our proposed  \kmethod is  better than the baseline Q-BERT,  under most settings of group numbers. 
 \paragraph{Larger group number will result in better performance.}
  Theoretically, the setting with a larger group number will result in better performance, as there will be more flexibility in weight assignments. This hypothesis is confirmed by the results shown in Table \ref{tab:group} that the performance significantly grows as the increase of group numbers. For example, by changing the group number from 1 to 128, the performance of both our \kmethod and baseline method increase by at least 2\%. 
   \paragraph{The trade-off between performance and complexity.}
   Although a larger group number will always bring better performance, such improvement is not without a cost, as the model have to infer more parameters.
   And as can be seen from the Table, the growth of improvement is ``slowed down'' as the number of groups increases.
   For example, there is at least 2\% improvement when increasing the number of groups from 1 to 128.
   However, only 0.1\% performance gain is obtained when we further increase the group numbers from 128 to 768.

\subsubsection{Combination with Knowledge Distillation}

Knowledge distillation methods are orthogonal to the present work.
The results of combining our method and knowledge distillation method DistilBERT are shown in Table \ref{tab:distil}. 
As can be seen from the table, DistilBERT will result in great performance loss even at a small compression ratio.
For example, in SST-2 dataset, the original DistilBERT brings more than $2\%$ performance drop in accuracy, but only reduces the size of base BERT in half. 
In contrast, after integrating with our \kmethod, the model is further compressed to $1/20$ of the original BERT, with only $0.1\%$ extra performance loss compared to using DistilBERT alone. 
This phenomenon indicates two practical conclusions:
\begin{packed_item}
    \item It is safe to integrate our \kmethod with knowledge distillation methods to achieve the extreme light-weight compact model.
    \item Compared to the method based on knowledge distillation, our proposed quantization method is more efficient, with a relatively higher compression ratio and a lower performance loss. 
\end{packed_item}

\section{Conclusion}
Recently, the compression of large Transformer-based models has attracted more and more research attentions. 
In this work, we proposed a two-level framework to achieve automatic mixed-precision quantization for BERT.
Under this framework, both the weights and precision assignments are updated through gradient-based optimization.
The evaluation results show that our proposed \kmethod is always better than baseline Q-BERT in all four NLP tasks, especially in the ultra-low bit setting.
Also, our \kmethod is orthogonal to the knowledge distillation solutions, which can together bring in extreme light-weight compact models with little performance loss.
These advantages make our method a practical solution for the resource-limited device (e.g., smartphone). 

\clearpage
\bibliographystyle{named}
\bibliography{references_abs}



\end{document}